\documentclass{article}
\usepackage[utf8]{inputenc}
\usepackage{biblatex}
\usepackage{graphicx}
\usepackage{scrextend}
\usepackage{algorithm}
\usepackage{amsmath}
\usepackage{algpseudocode}
\usepackage{hyperref}
\usepackage{authblk}
\hypersetup{
    colorlinks=true,
    linkcolor=blue,
    filecolor=magenta,      
    urlcolor=cyan,
    pdftitle={Overleaf Example},
    pdfpagemode=FullScreen,
    }

\title{CULT: Continual Unsupervised Learning with Typicality-Based Environment Detection}
\author[1]{Oliver Daniels-Koch}
\date{May 2022}
\affil[1]{Brandeis University}

\begin{document}

\maketitle

\begin{abstract}
    We introduce CULT (Continual Unsupervised Representation Learning with Typicality-Based Environment Detection), a new algorithm for continual unsupervised learning with variational auto-encoders. CULT uses a simple typicality metric in the latent space of a VAE to detect distributional shifts in the environment, which is used in conjunction with generative replay and an auxiliary environmental classifier to limit catastrophic forgetting in unsupervised representation learning. In our experiments, CULT significantly outperforms baseline continual unsupervised learning approaches. \footnote{code for this paper can be found here: https://github.com/oliveradk/cult}
\end{abstract}
\section{Introduction}
Deep neural networks often suffer from catastrophic forgetting - when learning a new task, their performance on previously learned tasks rapidly degrades. Continual learning is the problem of overcoming catastrophic forgetting. While much of the continual learning literature focused on supervised tasks, here we are interested in unsupervised continual learning, that is, learning a task-agnostic representations of multiple environments. When learning such task-agnostic representations, it is often useful to detect environmental shifts, which in turn requires some anomaly detection method. 
	Variational Autencoders \cite{kingma_auto-encoding_2013} provide one possible method. Since VAE's parameterize a likelihood function on the data, we can detect new environments (anomalies) by monitoring whether the likelihood function exceed some threshold. However, multiple studies \cite{nalisnick_deep_2019}, \cite{hendrycks_deep_2019}, \cite{choi_waic_2019} have found that likelihood is not a reliable metric for out-of-distribution detection - out of distribution samples will often be assigned a \textit{higher} likelihood than in-distribution samples. We produce similar results, finding that a VAE trained on FashionMNIST \cite{xiao_fashion-mnist_2017} assigns higher likelihood to out of distribution MNIST samples than to in distribution FashionMNIST samples. 
    As \cite{choi_waic_2019} notes, this defect of likelihood may be caused by the different between likelihood and \textit{typicality}. Typical samples (those that have information value close to the expected information) are different from the maximally likely samples, and while any given sample in the \textit{typical set} is less likely than the maximally likely sample, a sequence of samples from the typical set is more likely than a sequence of maximally likely samples. Inspired by \cite{achille_life-long_2018}, we use this insight for a new typicality-based environmental detection algorithm, monitoring the KL divergence between a normal distribution parameterized by the mean and standard deviation of a given batch of latent variables. This method reliably detects environmental shifts from FashionMNIST to MNIST in both directions. When coupled with generative replay \cite{shin_continual_2017}, atypicality based anomoly detectoin provides a simple, effective solution to unsupervised continual learning with variational auto encoders. 
\section{Related Work}
\subsection{Continual Learning}
Continual learning can roughly be split into three categories: rehearsal, gradient flow, and architecture expansion. Rehearsal (or replay) methods use past examples (either stored \cite{mnih_playing_2013}, or generated \cite{shin_continual_2017}) to train on while learning new tasks. Gradient based methods \cite{kirkpatrick_overcoming_2017}, \cite{zenke_continual_2017} avoid updating parameters were important for previous tasks. Architectural expansions dynamically allocate new model capacity for new tasks and environments \cite{rusu_progressive_2016} \cite{draelos_neurogenesis_2017}. 
\subsection{Continual Unsupervised Learning}
While much of the continual learning literature focused on continuously learning \textit{tasks}, like classification and game-playing, there is also work focused on learning representations continually on unlabeled data. CURL \cite{rao_continual_2019} used dynamic expansion on a VAE, allocating more capacity when the likelihood of samples falls below some threshold. VASE\cite{achille_life-long_2018} also relies on a variational auto-encoder, but uses generative replay and latent masking to prevent catastrophic forgetting, with latent masks determined by a typicality metric on each latent. We use a similar metric to detect new environments, as described below.  
\subsection{Anomaly Detection with Generative Models}
Intuitively, likelihood should be a useful metric for anomaly/out-of-distribution detection for probabilistic generative models. However, a mounting body of work \cite{nalisnick_deep_2019}, \cite{hendrycks_deep_2019}, \cite{choi_waic_2019} has shown that likelihood metrics often assign higher likelihoods to out of distribution samples. \cite{choi_waic_2019} hypothesizes that this failure is due to the difference between likelihood and typicality, but shows that their method is also does not promote typical points as more "in-distribution". We are aware of only one work (\cite{achille_life-long_2018}) that directly uses a typicality-based metric for unsupervised continual learning, but this metric was used in the context of latent masking, rather than "pure" anomaly detection for environment inference. 

\section{Typicality}
\subsection{Formal Definition}
For an i.i.d sequence $x_1, x_2, ..., x_n$ drawn from an alphabet $\mathcal{X}$, the typical set $T_{\epsilon}^{(n)}$ is given by all such sequences satisfying the following inequality: 

$$H(x) - \epsilon \leq - \frac{1}{n}\log{p(x_1, x_2, ..., x_n)} \leq H(X) + \epsilon$$

such that in the limit the average information of elements in the typical set converges to the entropy of the distribution.

Using this definition, we define the "most typical" set $T_0^{(n)}$ as all such sequences exactly satisfying the equality 

$$ -\frac{1}{n}\log{p(x_1, x_2, ..., x_n) }= H(x)$$

\subsection{Volume in High Dimensional Space}
To get an intuitive sense for why typicality-based metrics are favorable in comparison to likelihood, we need to review on how volume operates in high dimensional space. Following the work of \cite{betancourt_conceptual_2018}, for a neighborhood $\mathcal{N}$ in $\mathbf{R}^d$, the relative volume of $\mathcal{N}$ decreases as a function of $d$. When considering a density function $\pi$ defined on $R^d$, then as $d$ increases, most of the mass of $\pi$ will be concentrated where the density function \textit{and} the volume are sufficiently large. Notably, the mass does not necessarily concentrate at the mode. 

\subsection{The Typical Set of the Isotropic Gaussian}
This interplay between density and volume is made clear in the case of the multivariate isotropic gaussian, where the distance of the "most typical" samples to the mean is a monotonically increasing function of the dimensional of the sample space. Given $x \sim \mathcal{N}(\mathbf{0} \in R^d, \mathbf{I} \in R^{dxd})$, the most typical set is given by
$$T_0^{(n)}(\mathcal{N}(\mathbf{0}, \mathbf{1})) = \{\mathbf{x} | -\log{p(\mathbf{x})} = H[\mathcal{N}(\mathbf{0}, \mathbf{1})] \}$$ which reduces to 

$$T_0^{(n)}(\mathcal{N}(\mathbf{0}, \mathbf{1})) = \{\mathbf{x} | \lVert \mathbf{x} \rVert = \sqrt{d}\}$$See \ref{typ_deriv} for the computation. As the example above shows, drawing samples close to the mean of a multivariate gaussian becomes less likely as the dimensional of the sample space increases. So, if our distribution is approximated by a neural network, samples which are far from the most typical set (either close to the mean or very far from the mean), should be treated as anomalies. As stated below, we approximate this typicality "distance" using KL divergence from the isotropic gaussian prior.

\section{CULT}
\subsection{Problem Formalism}
We assume a distribution $s \sim Cat(\pi_1, ..., K)$ over a set of $K$ environments $\mathcal{S} = \{s_1, s_2, ..., s_K\}$, which share $N$ latent generative factors from the set$\mathcal{Z} = \{z_1, z_2, ..., z_N\}$. $\mathbf{z}$ is assumed to be standard normal, that is $\mathbf{z} \sim \mathcal{N}(0,1)$. A dataset $\mathbf{x}^s \sim p(\cdot | \mathbf{z}^s, s)$ can be synthesized from latents conditioned on the environment. Together, we have the following generative process:
$$ \mathbf{z} \sim \mathcal{N}(0,1) , \hspace{1cm} s \sim \textrm{Cat}(\pi_1, ..., K), \hspace{1cm} \mathbf{x}^s \sim p(\cdot |\mathbf{z}^s, s)$$
This setup is a simplified version of that in \cite{achille_life-long_2018}, doing away with latent masking. 
\subsection{Inferring the Generative Factors}
We use standard variational inference \cite{kingma_auto-encoding_2013} to infer the generative factors, maximizing the ELBO and using the re parameterization trick to efficiently compute gradients through the sampling of the normally distributed latents. This ELBO loss is given by: 
$$ \mathcal{L}_{ELBO} = E_{\mathbf{z}^s \sim q_{\phi}(\cdot|\mathbf{x}^s)}[-\log{p_{\theta}(\mathbf{x} | \mathbf{z}^s, s)] + D_{KL}(q_{\phi}(\mathbf{z}^s|\mathbf{x}^s)||p(\mathbf{z}))}$$
Note that unlike \cite{achille_life-long_2018}, we do not promote disentanglement by scaling the KL-Divergence term \cite{higgins_early_2016}, \cite{burgess_understanding_2018}. 

\subsection{Initializing and Inferring the Environment}
To detect new environments, we use a typicality metric taken over each batch. For each batch $\mathbf{x}^s_B$, we compute the latent features 
$\mathbf{z}^s_B$. We then compute the empirical mean and standard deviation of the latent batch, use these statistics to fit a diagonal gaussian, and take the KL Divergence of this gaussian and the standard normal distribution:
$\alpha = \sum_{i=1}^{N}{D_{KL}(\mathcal{N}(\bar{\mu}_n, \bar{\sigma}_n})||\mathcal{N}(0,1))$

We maintain an environment index $m$ and  a boolean $learning$, and define two hyperparameters $\lambda_0, \lambda_1$. If $alpha > \lambda_1$ and $learning$ is false, then we initialize a new environment, setting $learning$ to true and incrementing $m$. If $learning$ is true and $\alpha < \lambda_0$, then we set $learning$ to false. Else, we use an environment inference network $q_{\psi(s|\mathbf{x}_B)}$ to determine the environment $s_B$ of the batch. The intuition here is that we want to capture spikes in the atypically metric when entering a new environment, without allocating multiple indices to the same environment. See Algorithm 1 for the psuedo-code.
\begin{algorithm}
\caption{Typicality-Based Environment Detection and Inference}
\begin{algorithmic}
\State $\lambda_0, \lambda_1$
\State $m\gets 0$
\State $learning \gets True$
\For{$\mathbf{x}_B \hspace{0.1cm}\textrm{in} \hspace{0.1cm} \mathbf{X}$}
	\State $\hat{s} \gets q_{\psi}(\cdot|\mathbf{x}_B)$
   	\State $\mathbf{z}_B \gets q_{\phi}(\cdot | \mathbf{x}_B)$
    \State $\alpha \gets \sum_{i=1}^{N}{D_{KL}(\mathcal{N}(\bar{\mu}_n, \bar{\sigma}_n})||\mathcal{N}(0,1))$
    \If{$\alpha > \lambda_1 \hspace{0.1cm} \textrm{and} \hspace{0.1cm} learning = False$}
    	\State $m \gets m + 1$
        \State $learning \gets True$ 
        \State $s \gets m$ 
     \ElsIf{$\alpha < \lambda_0 \hspace{0.1cm} \textrm{and} \hspace{0.1cm} learning = True$}
        \State $learning \gets False$
        \State $s \gets \hat{s}$
   \Else{}
       \State $s \gets \hat{s}$
   \EndIf
        
\EndFor
\end{algorithmic}
\end{algorithm}

\subsection{Preventing Catastrophic Forgetting}
We use generative replay to mitigate catastrophic, leveraging the accurate environment detector by copying and freezing the current model after an environment is initialized. On each training iterations, "hallucinated" samples are generated from the frozen model via Monte-Carlo sampling. These samples are passed through the frozen model and current model, with a loss penalizing the encoder and decoder proximity of the current model to the frozen model on the samples.

$$\mathcal{L}_{past}(\phi, \theta) = E_{\mathbf{z}, s', \mathbf{x}'}
\big[D[q_{\phi}(\mathbf{z}|\mathbf{x}'), q_{\phi'}(\mathbf{z}'|\mathbf{x}')] + D[q_{\theta}(\mathbf{x}|\mathbf{z}, s'),  q_{\theta'}(\mathbf{x}'|\mathbf{z}, s')\big]$$
\subsection{Environment Inference Network}
The environment inference network is a linear layer added to the final layer of the encoder network, trained to predict the environment index. The training target on the current environment is given by the current environment index. To prevent catastrophic forgetting, the network is also trained on the hallucinated samples:

$$ \mathcal{L}_{env} = E_{\mathbf{x}}[-\log{q_{\phi}(\hat{s}|\mathbf{x})}] + E_{\hat{s} \neq s <m} E_{\mathbf{x}^{'} \sim p_{\theta'}(\mathbf{x}'|\mathbf{z}', s)}[-\log{q_{\phi}(s|\mathbf{x}')}]$$
\section{Experiments}
\subsection{Continual Unsupervised Learning with CULT}
We trained CULT on FashionMNIST to MNIST and MNIST to FashionMNIST for 10 epochs on each environment. The encoder and decoder were fully connected networks with one hidden layer of size 50, and a latent bottleneck of 16 variables. We set $\lambda_1=1.4$, $\lambda_0 = 0.25$. Adam was used with a learning rate of 1e-3 for the main network and 1e-4 for the environment network. The environment network also uses dropout with a dropout rate of .5. 

\subsection{Evaluating Catastrophic Forgetting}
To evaluate catastrophic forgetting, we measure the reconstruction loss of a trained model on a dataset seen earlier in training. To test how well the model has preserved representations useful for downstream tasks, we also train classifiers with 1 hidden layer of size 50 on the latent representations. 
    
\subsection{Baseline}
To compare the performance of CULT to relevant baselines, we ran FashionMNIST to MNIST while updating the replay network at a fixed interval of 500 steps, and FashionMNIST to MNIST with no generative replay.
\subsection{Results}
\subsubsection{Reconstruction Loss}
\begin{figure}[htp]
    \centering
    \includegraphics[width=6cm]{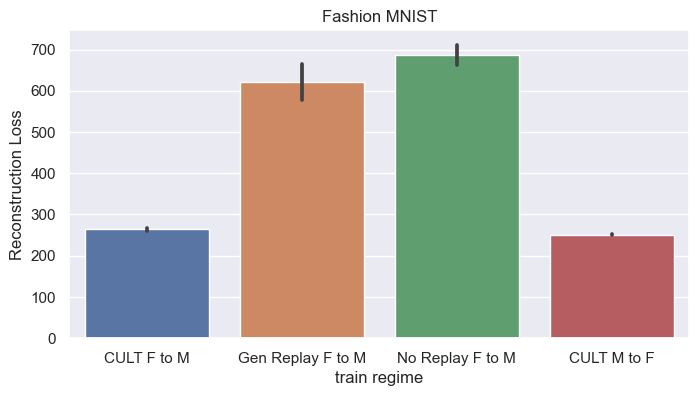}
    \includegraphics[width=6cm]{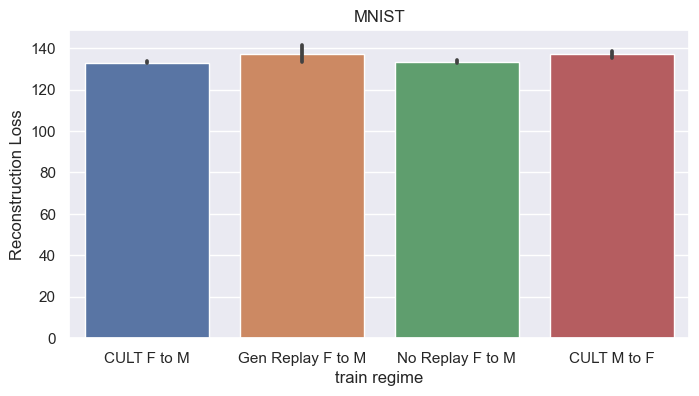}
    \caption{Reconstruction Losses}
     \label{fig:rec_losses}
\end{figure}

CULT dramatically outperforms baseline generative replay, more than halving reconstruction loss on FashionMNIST. In fact, the CULT model trained on FashionMNIST then MNIST has only \textit{slightly} higher reconstruction loss on FashionMNIST then the CULT model trained on MNIST then FashionMNIST. A similar result holds on MNIST reconstruction loss, with the MNIST to FashionMNIST CULT model achieving an only slightly higher reconstruction loss on MNIST then the FashionMNIST to MNIST CULT model (See figure \ref{fig:rec_losses}). 

\subsubsection{Latent Classification}
\begin{figure}[htp]
    \centering
    \includegraphics[width=12cm]{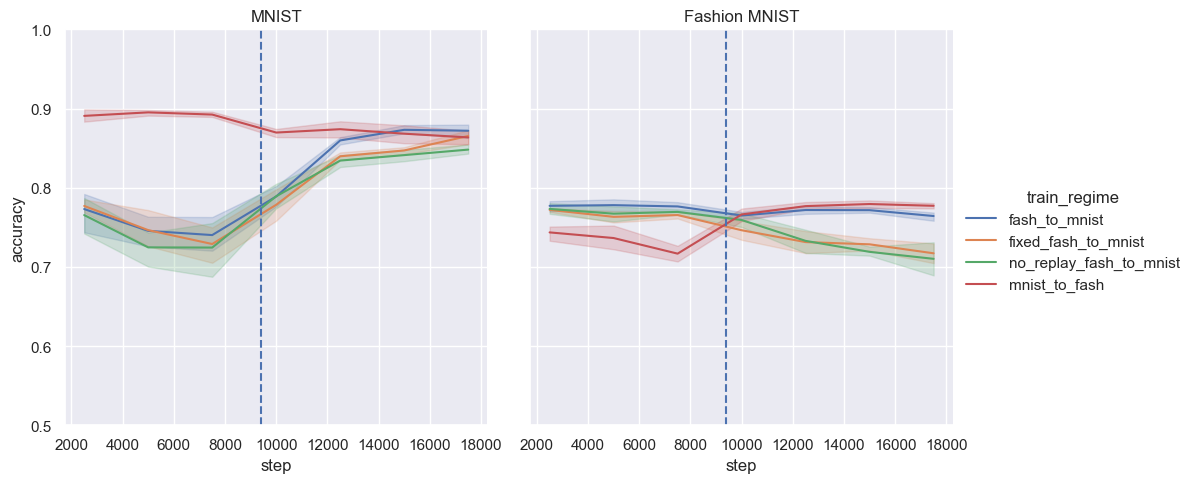}
    \caption{Classification Accuracy}
     \label{fig:class_acc}
\end{figure}
Classifiers trained on CULT's latent representations of FashionMNIST achieve a higher accuracy then classifiers trained on the latent representations produced by generative replay and no generative replay models. On MNIST, the CULT model trained on MNIST to FashionMNIST achieves rough parity with the other FashionMNIST to MNIST models, even out performing the no replay FashionMNIST to MNIST model.
Figure \ref{fig:class_acc} illustrates these training dynamics. Note how after the datasets switch (indicated by the blue dotted line), classification accuracy rapidly degrades for the generative replay and non generative replay models, but only slightly decreases and then stabilizes for the CULT FashionMNIST to MNIST model.

\subsubsection{Anomaly Detection Metrics}
\begin{figure}[htp]
    \centering
    \includegraphics[height=20cm]{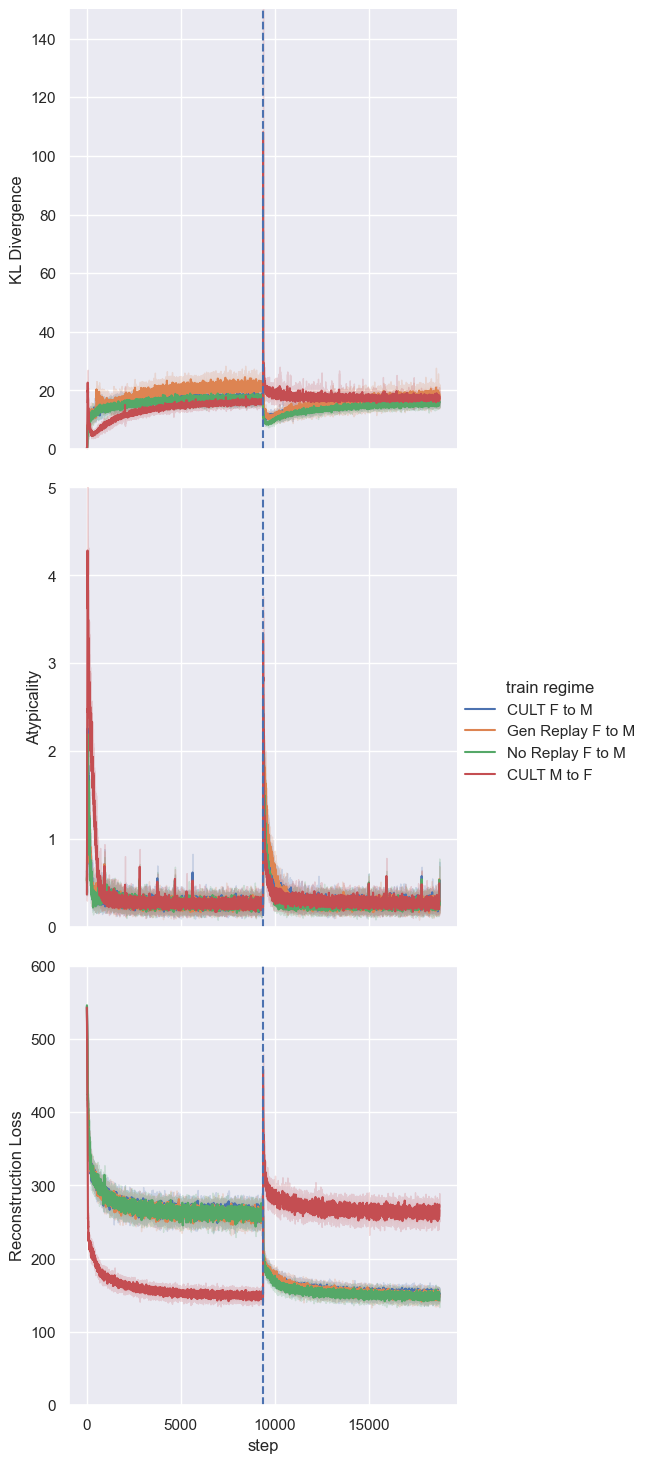}
    \caption{Anomaly Detection Metrics}
     \label{fig:anom_metrics}
\end{figure}
Tracing the KL divergence, reconstruction loss, and atypicality training curves in figure \ref{fig:anom_metrics} validates the hypothesis that typicality based anomaly detection methods overcome outstanding issues with likelihood based metrics. In particular, note that in the transition from FashionMNIST to MNIST (again indicated by the dotted blue line), the KL divergence and reconstruction loss \textit{decrease}, while the atypicality score \textit{increases}. MNIST samples are assigned a higher likelihood, but are atypical - thus using atypicality based anomaly detection is more successful than likelihood based anomaly detection. 

\section{Conclusion}
We presented CULT, a simple framework for using typicality-based environmental detection to facilitate continual learning with variational autoencoders and generative replay. Potential avenues for future work include running experiments on larger models and more complex datasets, experimenting with different typicality metrics, and performing a more rigorous analysis of typicality in generative probabilistic models. 

\printbibliography

\appendix
\section{Typicality Derivation}
\label{typ_deriv}
First note that the entropy (in nats) of a multivariate gaussian  $\mathcal{N}(\mu, \Sigma)$ is given by 
$$H[\mathcal{N}(\mu, \Sigma)] = \frac{1}{2} \ln{|\Sigma|} + \frac{D}{2}(1 + \ln{2\pi})$$
which, in the case of $\Sigma = \mathbf{I}$, reduced to 
$$H[\mathcal{N}(\mathbf{0}, \mathbf{I})] = \frac{D}{2}(1 + \ln{2\pi})$$
Next, we compute the negative log probability of a sample $x \sim \mathcal{N}(\mathbf{0}, \mathbf{I})$
\begin{align*}
	- \ln{p(x)} & = - \ln\left[(2\pi)^{-\frac{d}{2}} \Sigma^{-\frac{1}{2}} \exp(-\frac{1}{2} (x - \mu)^T \Sigma^{-1}(x-\mu))\right] \\ 
    & = -\ln\left[(2\pi)^{-\frac{d}{2}} \exp(-\frac{1}{2}\lVert x \rVert^2)\right] \\
    & = \frac{D}{2}\ln(2\pi) + \frac{1}{2} \lVert x \rVert^2
\end{align*}
Setting these expressions equal and canceling like terms, we have 

\begin{align*}
    \frac{1}{2} \lVert x \rVert^2 & = \frac{D}{2} \\
    \lVert x \rVert & = \sqrt{D}
\end{align*}

\end{document}